%% file: main.tex
\def\paperTitle{
SparseFlex: High-Resolution and Arbitrary-Topology 3D Shape Modeling
}

\def\authorBlock{
    Xianglong He$^{1,2\star}$ \qquad
    Zi-Xin Zou$^{2\star}$ \qquad
    Chia-Hao Chen$^{1,2}$ \qquad
    Yuan-Chen Guo$^{2}$ \qquad
    Ding Liang$^{2}$ \\
    Chun Yuan$^{1\dag}$ \qquad
    Wanli Ouyang$^{3}$ \qquad
    Yan-Pei Cao$^{2}$ \qquad
    Yangguang Li$^{2\dag}$ \\
    $^1${Tsinghua University} \qquad $^2${VAST} \qquad
    $^3${The Chinese University of Hong Kong} \\
}

\PassOptionsToPackage{dvipsnames,table}{xcolor}
\documentclass[10pt,twocolumn,letterpaper]{article}
\usepackage[pagenumbers]{cvpr}

\usepackage{bbm}
\usepackage{graphicx}	
\usepackage{amsmath}	
\usepackage{amssymb}
\usepackage{pifont}
\usepackage{booktabs}
\usepackage{times}
\usepackage{microtype}
\usepackage{epsfig}
\usepackage{caption}
\usepackage{float}
\usepackage{placeins}
\usepackage{color, colortbl}
\usepackage{stfloats}
\usepackage{enumitem}
\usepackage{tabularx}
\usepackage{xstring}
\usepackage{multirow}
\usepackage{xspace}
\usepackage{url}
\usepackage{subcaption}
\usepackage[hang,flushmargin]{footmisc}
\usepackage{cuted}
\usepackage{changepage}

% ...

\newcommand*{\Representation}{{SparseFlex}}

\usepackage{xr-hyper}

\makeatletter
\newcommand*{\addFileDependency}[1]{
  \typeout{(#1)}
  \@addtofilelist{#1}
  \IfFileExists{#1}{}{\typeout{No file #1.}}
}

\makeatother

\definecolor{cvprblue}{rgb}{0.21,0.49,0.74}
\usepackage[pagebackref,breaklinks,colorlinks,allcolors=cvprblue]{hyperref}
\usepackage[capitalize]{cleveref}
\crefname{section}{Sec.}{Secs.}
\crefname{table}{Table}{Tables}
\crefname{figure}{Fig.}{Figs.}

\begin{document}
\title{\paperTitle}
\author{\authorBlock}
\maketitle

\begin{strip}
    \centering
    \includegraphics[width=\textwidth]{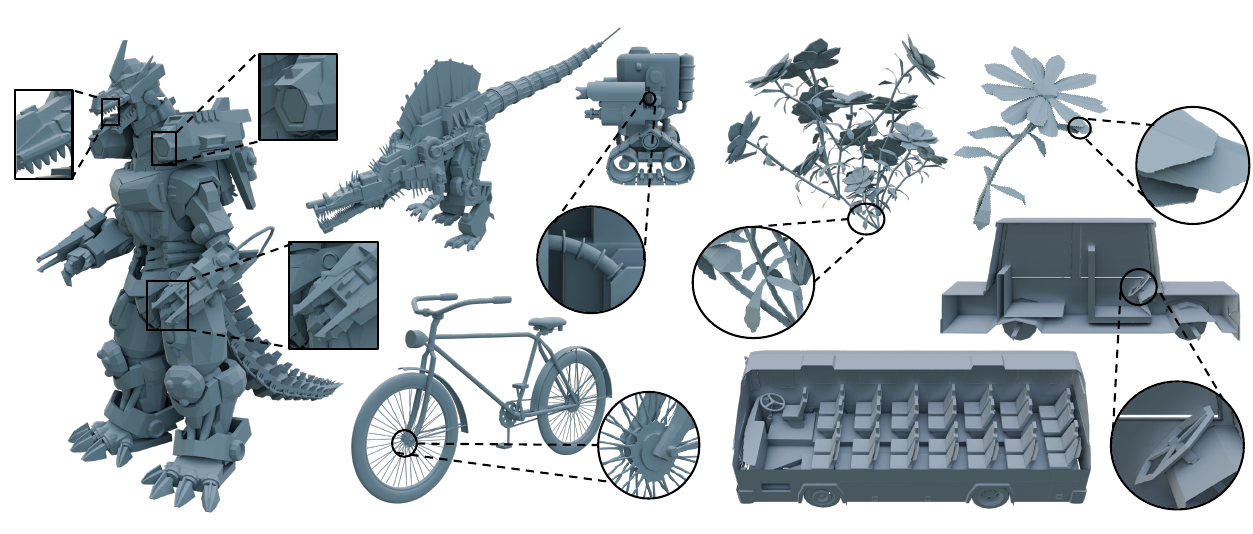}
    \vspace{-0.8cm}
    \captionof{figure}{\textbf{\Representation{} VAE achieves high-fidelity reconstruction and generalization from point clouds.} Benefiting from a sparse-structured differentiable isosurface surface representation and an efficient frustum-aware sectional voxel training strategy, our \Representation{} VAE demonstrates the state-of-the-art performance on complex geometries (left), open surfaces (top right), and even interior structures (bottom right), facilitating the high-quality image-to-3D generation with arbitrary topology.
    }
    \label{fig:teaser}
\end{strip}

\def\thefootnote{}\footnotetext{$\star$ Equal contribution. \\ $\dag$ Corresponding authors.} 
\input{00_abstract}
\input{01_intro}
\input{02_related}

\input{03_method}
\input{04_experiments}
\input{10_conclusion}

{\small
\bibliographystyle{ieeenat_fullname}
\bibliography{11_references}
}

\end{document}

%% file: 00_abstract.tex
\begin{abstract}
Creating high-fidelity 3D meshes with arbitrary topology, including open surfaces and complex interiors, remains a significant challenge. Existing implicit field methods often require costly and detail-degrading watertight conversion, while other approaches struggle with high resolutions. This paper introduces SparseFlex, a novel sparse-structured isosurface representation that enables differentiable mesh reconstruction at resolutions up to $1024^3$ directly from rendering losses. SparseFlex combines the accuracy of Flexicubes with a sparse voxel structure, focusing computation on surface-adjacent regions and efficiently handling open surfaces. Crucially, we introduce a frustum-aware sectional voxel training strategy that activates only relevant voxels during rendering, dramatically reducing memory consumption and enabling high-resolution training. This also allows, for the first time, the reconstruction of mesh interiors using only rendering supervision. Building upon this, we demonstrate a complete shape modeling pipeline by training a variational autoencoder (VAE) and a rectified flow transformer for high-quality 3D shape generation. Our experiments show state-of-the-art reconstruction accuracy, with a $\sim$82\% reduction in Chamfer Distance and a $\sim$88\% increase in F-score compared to previous methods, and demonstrate the generation of high-resolution, detailed 3D shapes with arbitrary topology. By enabling high-resolution, differentiable mesh reconstruction and generation with rendering losses, SparseFlex significantly advances the state-of-the-art in 3D shape representation and modeling. Please see our project page at \href{https://xianglonghe.github.io/TripoSF}{https://xianglonghe.github.io/TripoSF}.

\end{abstract}

%% file: 01_intro.tex
\section{Introduction}
\label{sec:intro}
3D Generative AI is rapidly advancing, with applications spanning entertainment, design, and robotics. However, generating 3D content is fundamentally more challenging than 2D image or text generation due to the inherent complexity of representing and manipulating 3D geometry. Achieving \emph{high fidelity} and supporting \emph{arbitrary topology}---including open surfaces and complex interiors---presents particularly significant hurdles.

Recent progress in 3D generative models has explored various representations, including point clouds~\cite{nichol2022point,melas2023pc2,vahdat2022lion,luo2021diffusion}, meshes~\cite{siddiqui2024meshgpt,nash2020polygen}, 3DGS~\cite{he2024gvgen,zhang2024gaussiancube,DBLP:journals/corr/abs-2412-01506,lan2024gaussiananything}, and implicit fields~\cite{zhang2024clay,DBLP:journals/tog/ZhangTNW23,ren2024xcube,li2025triposg,zhao2023michelangelo,cheng2023sdfusion,wu2024direct3d,li2024craftsman,DBLP:journals/corr/abs-2412-01506,zheng2023locally,chen20243dtopia}. Implicit field representations, such as Signed Distance Functions (SDFs) and occupancy fields, have shown favorable results~\cite{DBLP:journals/tog/ZhangTNW23,zhang2024lagem,wu2024direct3d,lan2024ln3diff,DBLP:journals/corr/abs-2412-01506,zheng2023locally}. However, creating training data for these methods typically involves a two-step process that limits their effectiveness. First, raw 3D mesh data must be converted into watertight representations~\cite{zhang2024clay,xu2014signed} to calculate SDF or occupancy values---a computationally expensive preprocessing step that often degrades fine details. Second, isosurfacing techniques (e.g., Marching Cubes~\cite{DBLP:conf/siggraph/LorensenC87}, Dual Contouring~\cite{ju2002dual}) are used to extract meshes from the learned continuous field, which can introduce further inaccuracies and artifacts. While Unsigned Distance Fields (UDFs) offer a potential way to model open surfaces~\cite{chibane2020neural,liu2023neudf,long2023neuraludf,DBLP:conf/eccv/YuDLLLLMKHTLW24,DBLP:conf/eccv/GuillardSF22,chen20223psdf}, they often suffer from instability and struggle to capture geometric fine details.

Rendering-based supervision offers a powerful, \emph{differentiable} alternative for training 3D representations and 3D generative models~\cite{wei2023neumanifold,shen2021deep,DBLP:journals/tog/ShenMHYWCGFSG23,DBLP:journals/corr/abs-2412-01506,xu2024instantmesh,wang2024crm}. By directly comparing rendered images of a generated mesh to ground-truth data, rendering losses avoid the need for the initial watertight preprocessing step and better preserve fine details. However, a critical bottleneck arises: when used with dense implicit fields, rendering supervision requires extremely high memory consumption at high resolutions, severely limiting the achievable fidelity.

This paper introduces \textbf{\Representation{}}, a new sparse-structured isosurface representation that addresses these limitations and unlocks high-resolution, differentiable mesh reconstruction and generation using rendering supervisions. \Representation{} is built upon Flexicubes~\cite{DBLP:journals/tog/ShenMHYWCGFSG23}, providing accurate and differentiable isosurface extraction. The key design is the use of a \emph{sparse} voxel structure instead of a conventional dense grid. This sparsity is crucial for two primary reasons: (1) it dramatically reduces memory consumption, enabling \emph{high-resolution modeling}, and (2) it allows for the effective pruning of voxels near open boundaries, enabling the representation of \emph{open surfaces}.

To fully leverage the capabilities of \Representation{}, we propose \textbf{frustum-aware sectional voxel training}. Inspired by techniques used in real-time rendering~\cite{akenine2019real}, this approach activates only the subset of \Representation{} voxels that reside within the camera's viewing frustum during each training iteration. We also introduce an adaptive strategy to control the frustum's parameters, further optimizing memory usage. This not only substantially reduces computational and memory overhead but also enables, for the first time, the reconstruction of mesh interiors using only rendering supervision by appropriately positioning the camera. \cref{fig:strategy} illustrates our \Representation{} representation and the efficient sectional voxel training strategy.

Building on \Representation{} and our frustum-aware training, we present a complete 3D shape modeling pipeline. We employ a variational autoencoder (VAE) architecture, drawing inspiration from TRELLIS~\cite{DBLP:journals/corr/abs-2412-01506} but with key modifications. Firstly, because our focus is on \emph{high-fidelity geometry}, we use point clouds as input to the VAE, providing a direct and detailed representation of the shape's surface. Furthermore, we introduce a \emph{self-pruning upsampling} module within the decoder to further refine the sparse voxel structure, which is particularly beneficial for representing open surfaces. A rectified flow transformer is then trained on the learned latent space for high-quality, image-conditioned 3D shape generation. Through extensive experiments on Toy4k~\cite{stojanov2021using}, ABO~\cite{collins2022abo}, GSO~\cite{downs2022google}, Meta~\cite{meta}, Objaverse~\cite{deitke2023objaverse}, and Deepfashion3D~\cite{zhu2020deep}, our method demonstrates state-of-the-art shape reconstruction accuracy with minimal detail degradation, and high-quality single-image 3D shape generation.

Our main contributions are:

\begin{itemize}
    \item We propose \Representation{}, a new sparse-structured isosurface representation enabling efficient, high-resolution, and differentiable 3D shape modeling, with natural handling of open surfaces.
    \item We introduce a novel sectional voxel training strategy with adaptive view frustum control, dramatically reducing memory consumption and enabling high-resolution mesh reconstruction and generation, including interiors, using rendering losses.
    \item We demonstrate state-of-the-art reconstruction accuracy and the generation of high-resolution, detailed 3D shapes with arbitrary topology, representing a significant advance in the field.
\end{itemize}

%% file: 02_related.tex
\section{Related Work}
\label{sec:related}

\subsection{3D Shape Representations for Generation}
\paragraph{Point Cloud.} Point clouds are a flexible 3D representation and can be easily acquired using depth sensors and LiDAR.
Research has focused on point cloud processing~\cite{qi2017pointnet,qi2017pointnet++,zhao2021point} and generation~\cite{nichol2022point,melas2023pc2,vahdat2022lion,luo2021diffusion}, with recent approaches treating point clouds as a distribution, directly sampling point clouds from noise via generative models~\cite{zhang2024geometry,chang20243d}.
However, due to their limitation in representing solid surfaces, an additional surface reconstruction step~\cite{kazhdan2013screened,huang2023neural,peng2021shape} is required.

\paragraph{Triangle Mesh.} Triangle meshes are the primary representation for 3D assets in modern industrial pipelines.
Recent works employ auto-regressive models to generate triangle faces sequentially~\cite{siddiqui2024meshgpt,nash2020polygen,tang2024edgerunner}, improving resemblance to artist-created meshes.
While effective for low-poly meshes, these methods struggle to produce high-quality meshes with a high face count.

\paragraph{Implicit Field.} Implicit fields (SDF or occupancy) are widely used in geometry learning, particularly for 3D reconstruction~\cite{ DBLP:journals/tog/ShenMHYWCGFSG23, DBLP:conf/eccv/GuillardSF22, wang2021neus,huang2023neural,zou2024gp,peng2020convolutional} and generation~\cite{zhang2024clay,DBLP:journals/tog/ZhangTNW23,ren2024xcube,li2025triposg,zhao2023michelangelo,cheng2023sdfusion,wu2024direct3d,li2024craftsman,DBLP:journals/corr/abs-2412-01506,zheng2023locally,chen20243dtopia}, producing high-quality meshes.
Enhancements with triplane~\cite{wu2024direct3d}, vector set~\cite{DBLP:journals/tog/ZhangTNW23,zhang2024lagem,zhao2023michelangelo} sparse voxels~\cite{DBLP:journals/corr/abs-2412-01506} and sparse hierarchical voxels~\cite{ren2024xcube} allow neural networks to decode the field more effectively. These methods typically use the isosurface techniques~\cite{DBLP:conf/siggraph/LorensenC87,nielson2004dual,ju2002dual} for surface extraction but struggle with open-surface shapes like clothing and flowers.
Additionally, many require a time-consuming watertight conversion that degrades details, while rendering supervised approaches~\cite{DBLP:journals/corr/abs-2412-01506} suffer from high memory consumption when training at high resolutions.
\input{figs/pipeline}
\paragraph{Open Surfaces.}
Meshes with open surfaces are common but remain challenging to process.
The Unsigned Distance Field (UDF) is widely used for open-surface modeling from point clouds~\cite{chibane2020neural,chen2022neural,ye2022gifs,wang2022hsdf} or multiview images~\cite{liu2023neudf,long2023neuraludf,meng2023neat}.
Surf-D~\cite{DBLP:conf/eccv/YuDLLLLMKHTLW24} introduces a UDF-based diffusion model for generating meshes with arbitrary topology. 
However, UDF-based shape modeling is more difficult than SDF, and its surface extraction~\cite{guillard2022meshudf} is prone to instability due to inaccuracy in neural network gradient estimations.
As a result, achieving high-quality open-surface meshes remains challenging.
3PSDF~\cite{chen20223psdf} introduces a three-pole sign to distinguish surface-adjacent regions but relies on binary occupancy grids for surface extraction, leading to discontinuities and artifacts on the surface.
Although G-Shell~\cite{DBLP:conf/iclr/LiuFXLPBS24} proposes a new 3D representation for extracting non-watertight meshes from watertight triangular meshes and trains a diffusion model based on this representation, its dense grid structure limits it to high resolution when handling complex shapes.
In this paper, we incorporate sparse structure into Flexicubes~\cite{DBLP:journals/tog/ShenMHYWCGFSG23} to efficient high-quality isosurface representation for open faces.

\subsection{3D Generative Models and VAE}
Existing 3D generation studies are primarily classified into two categories: large 3D reconstruction model following 2D multiview diffusion models and native 3D generation models.
The first category uses multiview diffusion models to generate multiview images from text or an image~\cite{DBLP:conf/iclr/LiuLZLLKW24,DBLP:conf/iclr/ShiWYMLY24,huang2024epidiff,wang2023imagedream,huang2024mv, liu2024unidream}, followed by a large 3D reconstruction model~\cite{DBLP:conf/iclr/Hong0GBZLLSB024,DBLP:conf/iclr/Li00XLXHSSB24,zou2024triplane,zhang2024gs,xu2024grm,xu2024instantmesh,tang2024lgm, tochilkin2024triposr} is employed to reconstruct the 3D representation from these images in seconds.
However, inconsistencies between generated multiview images significantly often degrade the result quality.
The second category focuses on native 3D generation model that directly generate 3D models through generative models, including GAN~\cite{wu2016learning,chan2022efficient,chen2019learning,zheng2022sdf}, auto-regressive~\cite{mittal2022autosdf,zhang20223dilg}, diffusion~\cite{vahdat2022lion,hui2022neural,DBLP:journals/tog/ZhangTNW23,zhang2024clay} and rectified flow~\cite{li2025triposg,DBLP:journals/corr/abs-2412-01506}.

Due to the diverse and non-compact nature of 3D representation, many approaches use Variational Auto-Encoder (VAE)~\cite{kingma2013auto} or Vector Quantized VAE (VQ-VAE)~\cite{van2017neural} to encode 3D shapes into latent spaces for the generative models.
Geometry-focused methods often input point clouds uniformly sampled from mesh surfaces into the VAE~\cite{DBLP:journals/tog/ZhangTNW23,chen2024dora}.
When aim at decoding both geometry and texture, some methods~\cite{DBLP:journals/corr/abs-2412-01506,lan2024ln3diff,lan2024gaussiananything} encode multiview image features into latent spaces.
This two-stage process makes the VAE's reconstruction quality crucial for subsequent generation performance. Some works~\cite{zhang2024lagem,chen2024dora} have improved generation quality by enhancing the VAE's shape encoding-decoding capabilities.
In this paper, we aim to develop a foundational VAE that encodes 3D shapes into latent spaces and reconstructs them with arbitrary topology while preserving the raw 3D shape's details.

%% file: figs/pipeline.tex
\begin{figure*}[ht]
    \centering
    \vspace{-0.7cm}
    \includegraphics[width=0.95\linewidth]{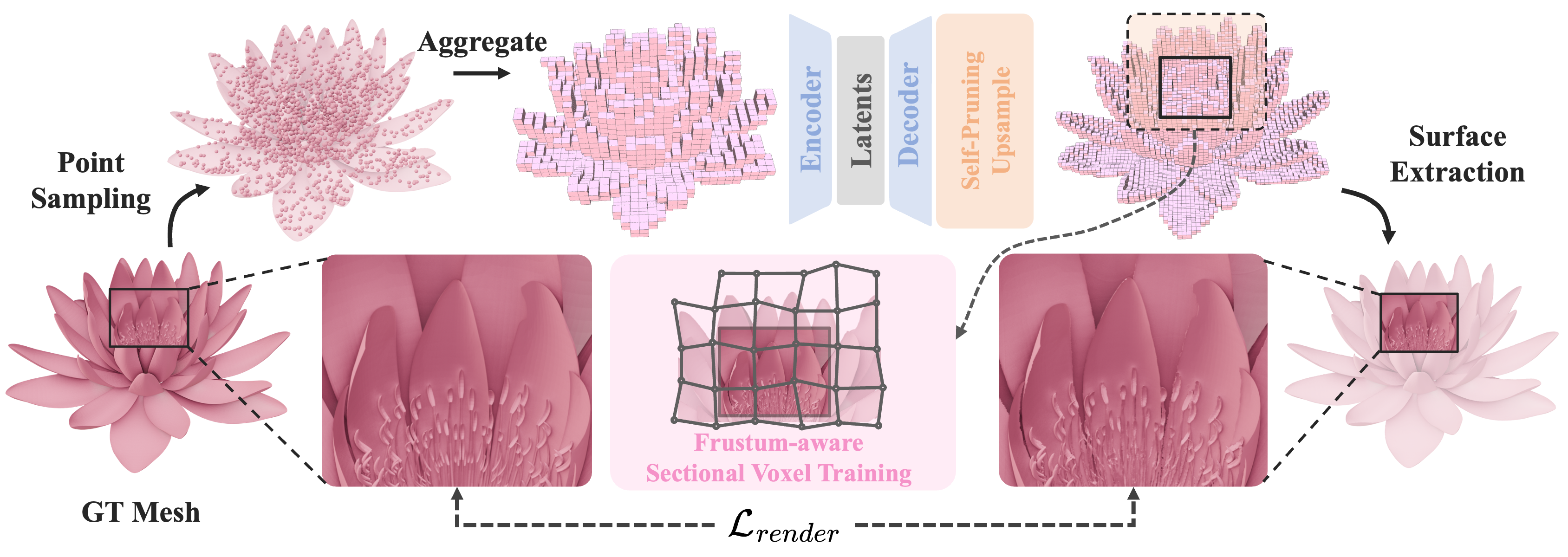}
    \vspace{-0.2cm}
    \caption{\textbf{Overview of the \Representation{} VAE pipeline.} \Representation{} VAE takes point clouds sampled from a mesh as input, voxelizes them, and aggregates their features into each voxel. A sparse transformer encoder-decoder compresses the structured feature into a more compact latent space, followed by a self-pruning upsampling for higher resolution. Finally, the structured features are decoded to \Representation{} through a linear layer. Using the frustum-aware section voxel training strategy, we can train the entire pipeline more efficiently by rendering loss.
    }
    \label{fig:pipeline}
    \vspace{-0.4cm}
\end{figure*}

%% file: 03_method.tex
\vspace{-0.2cm}
\section{Method}
\label{sec:method}

We present a method for high-resolution 3D shape modeling based on a novel sparse-structured isosurface representation, \Representation{}. Our approach leverages differentiable rendering for training, enabling accurate reconstruction of complex geometries, including open surfaces and interiors (see~\cref{fig:teaser}. ~\cref{fig:pipeline} illustrates our variational autoencoder (VAE) based shape modeling pipeline, which utilizes the \Representation{} representation to learn a compact latent space of 3D shapes. The VAE decoder outputs the parameters of a \Representation{} instance, facilitating high-resolution mesh reconstruction. A key component of our approach is \emph{frustum-aware sectional voxel training}, which significantly reduces memory consumption, allowing for training at resolutions up to $1024^3$. Details of the \Representation{} representation are provided in ~\cref{sec:representation}, followed by a description of the VAE architecture in ~\cref{sec:vae-model} and the training procedure in~\cref{sec:vae-training}. \cref{sec:gen} describes image-conditioned 3D shape generation using a rectified flow transformer on the learned latent space.

\input{figs/strategy}
\subsection{\Representation{} Representation}\label{sec:representation}

\vspace{-0.1cm}
\paragraph{Preliminary.} To achieve differentiable mesh extraction while preserving sharp features, we build upon Flexicubes~\cite{DBLP:journals/tog/ShenMHYWCGFSG23}, a method based on Dual Marching Cubes~\cite{nielson2004dual} (DMC). DMC places vertices at the center of voxels rather than edges/corners, leading to better feature preservation. 
Flexicubes constructs a \emph{dense} voxel with a resolution of $N_r^3$, where an SDF scalar grid $s \in \mathbb{R}^{N_g^3}$ is assigned to the voxel corner points, where $N_g=N_r+1$.
For each voxel, it incoporates interpolation weights $\alpha\in \mathbb{R}_{>0}^{N_r^3\times 8}, \beta\in \mathbb{R}_{>0}^{N_r^3\times 12}$ to each voxel cell, and deformation vectors $\delta\in\mathbb{R}^{N_g^3}$ to each SDF grid.
The underlying surface mesh can be effectively optimized via differentiable rendering~\cite{Laine2020diffrast}.

\vspace{-0.4cm}
\paragraph{Sparse Structured Flexicubes.}
The core design of \Representation{} is the introduction of a \emph{sparse} voxel structure, enabling high-resolution shape representation while drastically reducing memory consumption. Instead of a dense grid, \Representation{} represents a shape using a significantly smaller set of voxels, $\mathcal{V}$, concentrated near the surface. This sparsity is crucial for two reasons: (1) it allows us to achieve much higher resolutions than would be possible with a dense grid, and (2) it enables the natural representation of open surfaces by simply omitting voxels in empty regions.

Specifically, the \Representation{} is defined by a set of $N_v$ voxels, $\mathcal{V} = \{v_i = (x_i, y_i, z_i)\}$, where $v_i$ represents the 3D coordinates of the \emph{center} of the $i$-th voxel.  Let $N_c$ be the number of corner grids associated with these voxels, where $N_v = |\mathcal{V}|$.
Each voxel is associated with interpolation weights $\{\alpha_i\in \mathbb{R}_{>0}^{8}, \beta_i\in \mathbb{R}_{>0}^{12}| 0\le i < N_v\}$. Each corner grid is associated with an SDF value $\{s_j|0\le j<N_c\}$ and deformation vectors $\{\delta_j|0\le j<N_c\}$. Due to the sparsity, $N_v \ll N_r^3$ and $N_c \ll N_g^3$, representing a significant reduction in memory usage compared to the dense Flexicubes representation. We only apply Dual Marching Cubes on these sparse voxels to extract the underlying surface.
Formally, the \Representation{} representation, $\mathcal{S}$, is defined as:
{\small\begin{equation}
\addtolength{\abovedisplayskip}{-5pt}
\addtolength{\belowdisplayskip}{-5pt}
    \small\mathcal{S}=(\mathcal{V},\mathcal{F}_c,\mathcal{F}_v), \quad \mathcal{F}_c=\{s_j,\delta_j\}, \quad \mathcal{F}_v=\{\alpha_i,\beta_i\},
\end{equation}}
where $\mathcal{V}$ represents the voxel centers, $\mathcal{F}_c$ contains the SDF values and deformations at the corner grids, and $\mathcal{F}_v$ contains the interpolation weights for each voxel.

\Representation{} inherits the differentiability of Flexicubes, allowing for end-to-end optimization using rendering losses. This eliminates the need for watertight mesh pre-processing, preserving fine details. Furthermore, the sparse structure, combined with the continuous and deformable nature of the SDF, allows for accurate and efficient representation of high-quality open-surface meshes. The sparsity also paves the way for our efficient frustum-aware training strategy, described in~\cref{sec:vae-training}.

\subsection{SparseFlex VAE for Shape Modeling} \label{sec:vae-model}
To learn a compact and disentangled latent space of 3D shapes, we employ a variational autoencoder (VAE)~\cite{kingma2013auto} that utilizes the SparseFlex representation. A VAE learns a probabilistic mapping between an input space (in our case, 3D shapes represented as point clouds) and a lower-dimensional latent space, enabling both reconstruction and generation of shapes.~\cref{fig:pipeline} also illustrates our VAE architecture.  Our architecture draws inspiration from TRELLIS~\cite{DBLP:journals/corr/abs-2412-01506}, but with key modifications to leverage the strengths of SparseFlex.

\vspace{-0.4cm}

\paragraph{Encoder.} The input to our encoder is a point cloud $\mathcal{P} = \{p_i \in \mathbb{R}^3\}_{i=1}^{N_p}$, uniformly sampled from the surface of a 3D mesh, along with corresponding normals $\mathcal{N} = \{n_i \in \mathbb{R}^3\}_{i=1}^{N_p}$.  We first voxelize the point cloud to obtain the sparse structure $\mathcal{V}$ of the SparseFlex representation $\mathcal{S}$.  We then employ a shallow PointNet~\cite{qi2017pointnet} to aggregate local geometric features \emph{within each voxel}.  Specifically, for each voxel $v_i \in \mathcal{V}$, we apply a local max-pooling operation~\cite{peng2020convolutional} to the points contained within that voxel, producing a feature vector $f_i$. These voxel features $\mathcal{F} = \{f_i\}$, along with the sparse structure $\mathcal{V}$, are then fed into a sparse transformer backbone. This backbone utilizes shifted window attention~\cite{liu2021swin,yang2023swin3d}, similar to TRELLIS~\cite{DBLP:journals/corr/abs-2412-01506}, but is adapted to operate directly on the sparse voxel features $\mathcal{F}$ and structure $\mathcal{V}$. The transformer outputs a latent code $z \in \mathbb{R}^{d_z}$, which represents the encoded 3D shape.

\vspace{-0.4cm}

\paragraph{Decoder.} The decoder takes the latent code $z$ as input and predicts the parameters of a SparseFlex instance, $\mathcal{S} = (\mathcal{V}, \mathcal{F}_c, \mathcal{F}_v)$.  We use a series of transformer layers, culminating in a final linear layer, to predict the SDF values ($s_j$) and deformations ($\delta_j$) for each corner grid, as well as the interpolation weights ($\alpha_i$, $\beta_i$) for each voxel.

\vspace{-0.4cm}

\paragraph{Upsampling Modules.} To achieve high-resolution reconstructions, we incorporate two convolutional,  self-pruning upsampling modules within the decoder, following the transformer. These modules progressively increase the resolution of the SparseFlex representation. Each upsampling module \emph{subdivides} existing voxels into smaller voxels (increasing the resolution by a factor of $4$). Crucially, each module also $prunes$ redundant voxels based on a predicted occupancy value.  A voxel is considered occupied if, after the subdivision, it contains any points from the input point cloud $\mathcal{P}$. This pruning process, inspired by~\cite{ren2024xcube}, is essential for maintaining the sparsity of the SparseFlex representation and is particularly beneficial for accurately representing open surfaces, as it removes unnecessary voxels in empty regions.

\subsection{Training SparseFlex VAE}\label{sec:vae-training}
We train our SparseFlex VAE end-to-end using rendering losses, leveraging the differentiability of the SparseFlex representation and a new training strategy called \emph{frustum-aware sectional voxel training}. This strategy dramatically reduces memory consumption during training, enabling us to achieve high resolutions (up to $1024^3$) that would be infeasible with traditional approaches.

\vspace{-0.4cm}

\paragraph{Frustum-aware Sectional Voxel Training.} Even with the sparse structure of SparseFlex, directly rendering the entire representation at high resolutions can be computationally expensive. Furthermore, standard rendering supervision typically focuses only on the visible surface, neglecting the interior of the shape. Besides, recent methods~\cite{xu2024instantmesh,DBLP:journals/corr/abs-2412-01506,wei2024meshlrm,wang2024crm,liu2025meshformer} relying on rendering supervision from mesh typically require extracting the entire mesh because a dense representation doesn't trivially allow for partial extraction. In contrast, our sparse representation naturally enables partial extraction. To address these issues, we introduce frustum-aware sectional voxel training. Inspired by techniques used in real-time rendering for efficient visibility culling~\cite{akenine2019real}, this approach activates only the voxels within the camera's viewing frustum during each training iteration. \emph{``Activating''} a voxel means including it in the isosurface extraction and rendering process. This \emph{``sectional''} approach means we only process a \emph{portion} of the 3D space at a time, markedly reducing memory usage.

As illustrated in~\cref{fig:strategy}, given the camera's extrinsic $\pi$, intrinsics $K$, and the near ($n$) and far ($f$) clipping planes of the viewing frustum, we compute the Model-View-Projection (MVP) matrix $\mathbf{MVP}$. We then use a boolean operator to check whether the center of each voxel $v_i$ lies within the viewing frustum defined by the MVP matrix.  We use $I(v_i \in \text{Frustum}(\mathbf{MVP}))$ to represent this check, where $I(\cdot)$ is an indicator function. The set of active voxels, $\mathcal{V}_{\text{active}}$, is then defined as:
\begin{equation}
\small
\mathcal{V}_{\text{active}} = \{v_i | I(v_i \in \text{Frustum}(\mathbf{MVP})) = 1, v_i \in \mathcal{V} \}.
\end{equation}

\vspace{-0.6cm}

\paragraph{Adaptive Frustum and Interior Reconstruction.} We introduce a visibility ratio $\alpha$ ($0 < \alpha \leq 1$) controlling the \emph{proportion} of active voxels in SparseFlex. We adaptively adjust the near and far clipping planes to ensure that approximately $\alpha N_v$ voxels are within the frustum. This is achieved through an iterative process: we initially set the near and far planes and iteratively adjust them based on the number of active voxels until the desired proportion is reached.

This adaptive frustum also enables a novel capability: reconstructing mesh \emph{interiors} using only rendering supervision. By positioning a virtual camera \emph{inside} the object or adjusting the near clipping plane to intersect the mesh, we can render and supervise the internal structure (see~\cref{fig:performance_comparison}). Moreover, the zoom-in camera viewpoint can render the mesh surface with greater details, making it better for higher-resolution training. This is a significant advantage over methods that rely on watertight representations, which cannot capture interior details.

\vspace{-0.3cm}

\paragraph{Loss Function.}  We train our VAE in an end-to-end manner, with an objective function comprising four components:
\begin{equation}
\small
    \mathcal{L}=\lambda_1 \mathcal{L_{\text{render}}} + \lambda_2 \mathcal{L_{\text{prune}}} + \lambda_3 \mathcal{L_{\text{KL}}} + \lambda_4 \mathcal{L_{\text{flex}}},
\end{equation}
where $\lambda_1$, $\lambda_2$, $\lambda_3$, and $\lambda_4$ are weighting coefficients that balance the different loss terms.

$\mathcal{L}_{\text{render}}$ is the rendering supervision loss. We employ a combination of losses commonly used in differentiable rendering~\cite{mildenhall2021nerf,kerbl20233d}:
\begin{equation}
\small
\mathcal{L}_{\text{render}} = \lambda_d \mathcal{L}_d + \lambda_n \mathcal{L}_n + \lambda_m \mathcal{L}_m + \lambda_{ss} \mathcal{L}_{ss} + \lambda_{lp} \mathcal{L}_{lp},
\end{equation}
where $\mathcal{L}_d$, $\mathcal{L}_n$, and $\mathcal{L}_m$ denote the L1 loss for depth maps, normal maps, and mask maps, respectively. $\mathcal{L}_{ss}$ and  $\mathcal{L}_{lp}$ denote SSIM loss and LPIPS loss, and are only applied on normal maps.  Please refer to the supplementary material for a detailed definition of these losses and their weighting coefficients.

$\mathcal{L}_{\text{prune}}$ is the structure loss, formulated as a binary cross-entropy (BCE) loss, supervising the construction of sparse voxels:
\begin{equation}
\small
\mathcal{L}_\text{prune}=\text{BCE}\left(V,\hat{V}\right),
\end{equation}
where $V$ represents the ground-truth occupancy of voxels derived from the input point cloud and $\hat{V}$ represents the predicted occupancy by the upsampling modules.

$\mathcal{L}_{\text{KL}}$ is the KL divergence between the learned latent distribution and a standard normal prior, regularizing the latent space. $\mathcal{L}_{\text{flex}}$ is the regularization term from Flexicubes~\cite{DBLP:journals/tog/ShenMHYWCGFSG23} to encourage smooth SDF values.

\vspace{-0.1cm}
\subsection{Image-to-3D Generation with Rectified Flow}\label{sec:gen}

Building upon the trained \Representation{} VAE, we develop a pipeline for high-quality, image-conditioned 3D shape generation, following a similar approach to TRELLIS~\cite{DBLP:journals/tog/ShenMHYWCGFSG23}. Our approach consists of two main components: \textit{Structure Flow Model} and \textit{Structured Latent Flow Model}.

\vspace{-0.4cm}

\paragraph{Structure Flow Model.}
First, a separate, simple fully 3D convolutional structure VAE is employed to compress dense voxels representing 3D shapes into a low-resolution ($1/4$ scale) space. Subsequently, the image condition features are extracted using DINOv2~\cite{oquab2023dinov2} and injected into the transformer model via cross-attention, after which a rectified flow model is trained within this low-resolution space. During inference, given an input image, the trained structure flow model generates the corresponding low-resolution 3D space, which is then decoded by the structure VAE to produce the sparse structure of the generated 3D shape.

\vspace{-0.4cm}

\paragraph{Structured Latent Flow Model.}
Based on the proposed \Representation{} VAE, the point cloud and the corresponding voxelized sparse structure of a 3D shape are encoded into a \emph{structured latent space}. Subsequently, the image condition feature obtained via DINOv2 is injected into the sparse transformer model through cross-attention, followed by training a rectified-flow model within this structured latent space. During inference, given an input image, the corresponding sparse structure is first generated using the structure flow model and its structure VAE. Then, both the sparse structure and the input image are provided to the structured latent flow model to generate the corresponding structured latent representation. Finally, the \Representation{} VAE decodes this latent representation to produce the final 3D shape.

%% file: figs/strategy.tex
\begin{figure}[tp]
    \centering
    \vspace{-0.6cm}
    \includegraphics[width=1\linewidth]{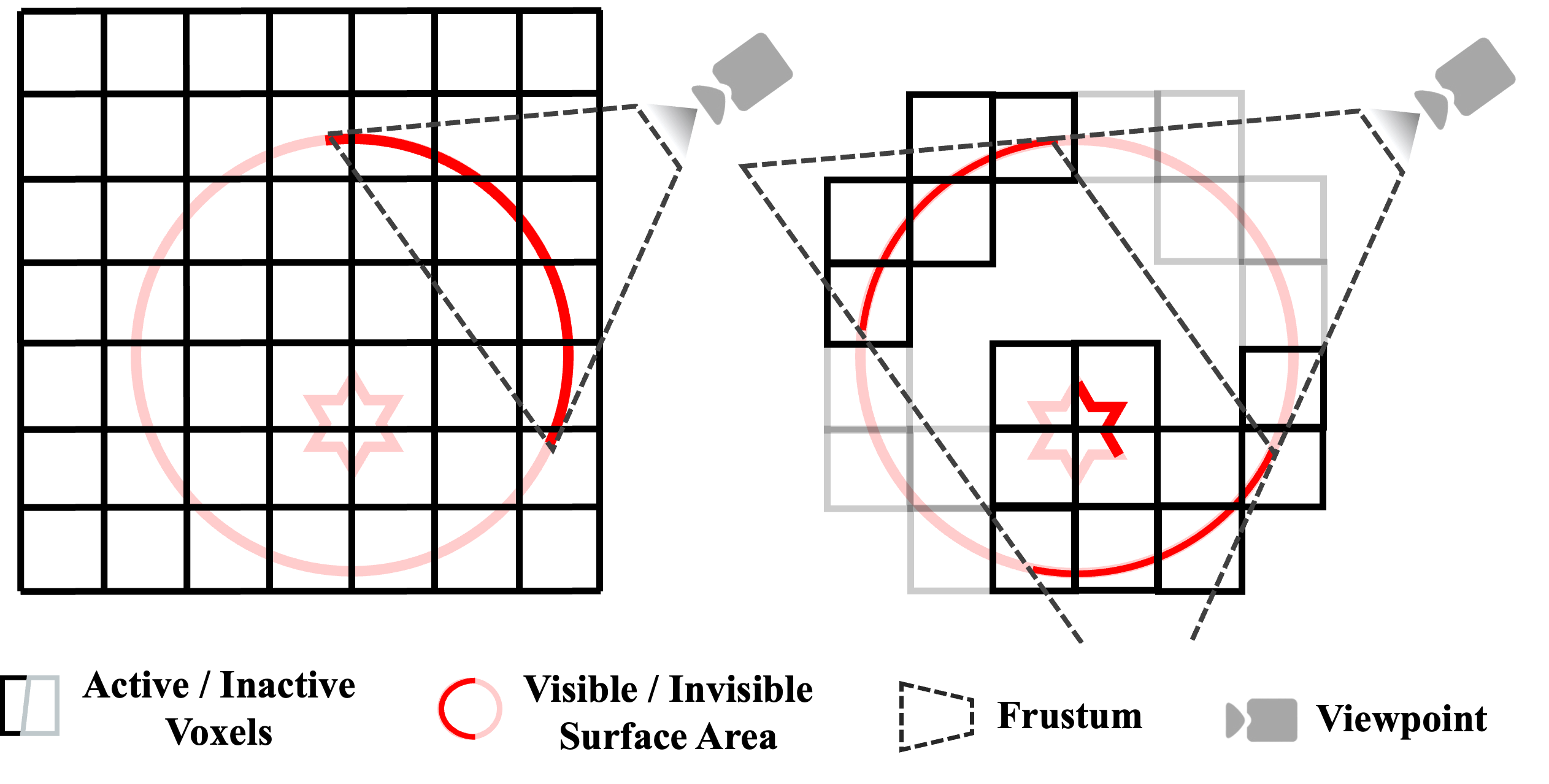}
    \vspace{-0.7cm}
    \caption{\textbf{Frustum-aware sectional voxel training.} The previous mesh-based rendering training strategy (left) requires activating the entire dense grid to extract the mesh surface, even though only a few voxels are necessary during rendering. In contrast, our approach (right) adaptively activates the relevant voxels and enables the reconstruction of mesh interiors only using rendering supervision.}
    \label{fig:strategy}
    \vspace{-0.5cm}
\end{figure}

%% file: 04_experiments.tex
\input{tables/vae_main}
\input{tables/deepfashion3d_main}

\section{Experiments}\label{sec:experiments}
\subsection{Implementation Details}
We develop the implementation of \Representation{} based on the official code$^{1}$ provided by FlexiCubes~\cite{DBLP:journals/tog/ShenMHYWCGFSG23}.
We train \Representation{} VAE and structured latent flow model using approximately 400K high-quality 3D meshes filtered from large-scale datasets, Objaverse (-XL)~\cite{deitke2023objaverse,deitke2023objaversexl}.
Since incorrect normals in raw data can significantly degrade the performance of both VAE reconstruction and image-to-3D generation, we apply a mesh preprocessing step to correct these flipped normals, ensuring that all normals are consistently oriented outwards. Please refer to the supplementary for more details.

Building on the success of progressive training in recent works~\cite{zhang2024clay}, we train \Representation{} VAE progressively, increasing final resolution from low to high (256, 512, and 1024).
For the structure VAE and structure flow model, we adopt the model from Trellis~\cite{DBLP:journals/corr/abs-2412-01506} and finetune their pre-trained weights to our task.
We train \Representation{} VAE on 64 A100 GPUs with a batch size of 64 and train structured latent flow models with a batch size of 256. We use the AdamW~\cite{loshchilov2017decoupled} optimizer with an initial learning rate of $1e-4$ and the weight decay as $0.01$.
At inference, we generate the results with 3.5 CFG and 50 sampling steps.
\def\thefootnote{}\footnotetext{$^1$\url{https://github.com/nv-tlabs/FlexiCubes}}

\begin{figure*}
    \vspace{-0.7cm}
    \centering
    \includegraphics[width=0.99\linewidth]{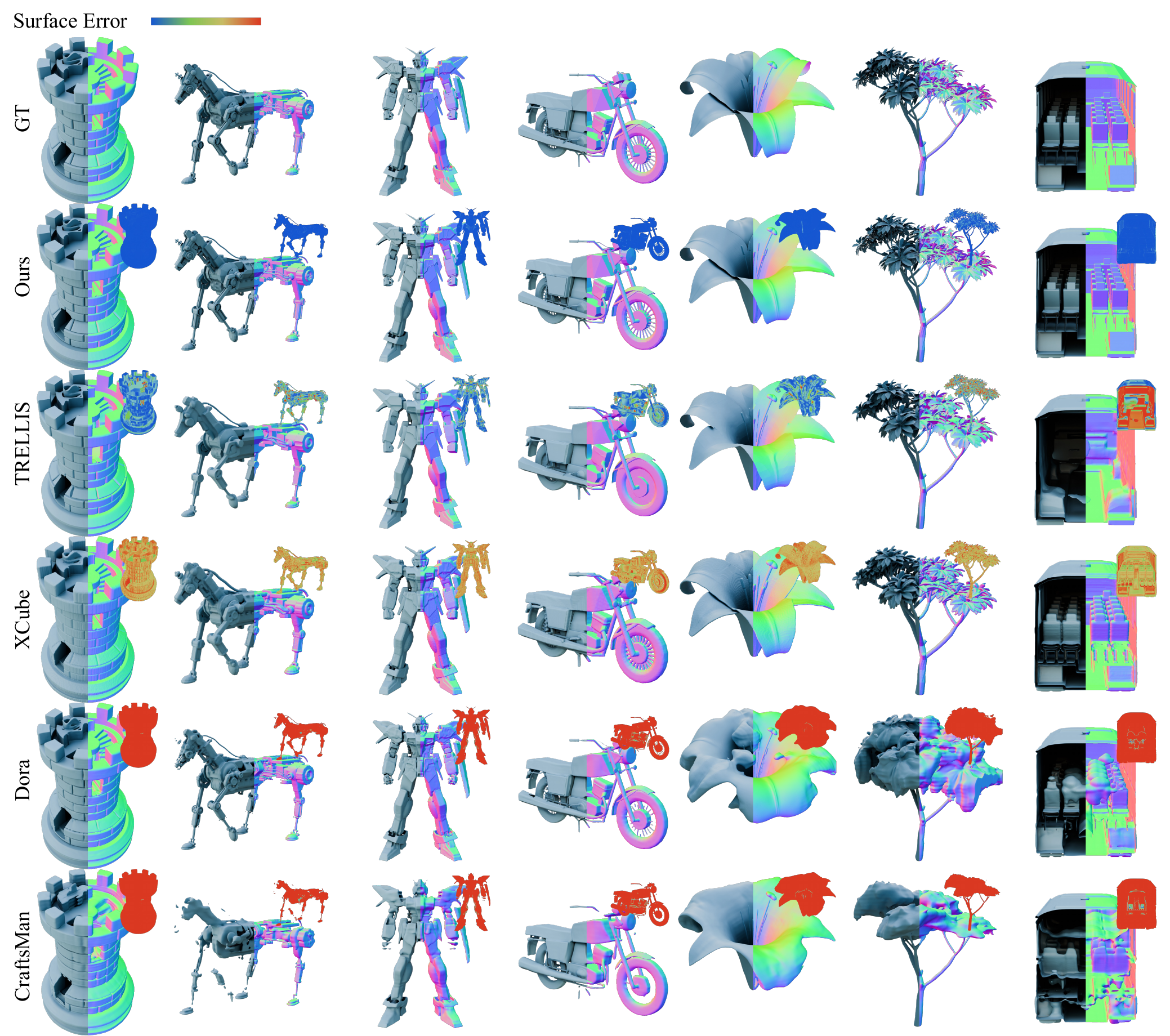}
    \vspace{-0.3cm}
    \caption{\textbf{Qualitative comparison of VAE reconstruction between ours and other state-of-the-art baselines.} Our approach demonstrate superior performance in reconstructing complex shapes, open surfaces, and even interior structures.}
    \vspace{-0.4cm}
    \label{fig:performance_comparison}
    
\end{figure*}

\subsection{Dataset, Baselines, and Metrics}
\paragraph{Dataset.}
We evaluate the reconstruction quality of VAE across different methods on a diverse set of datasets, including 1) universal datasets ABO~\cite{collins2022abo}, GSO~\cite{downs2022google}, Meta~\cite{meta}, Objaverse~\cite{deitke2023objaverse}, Toys4k~\cite{stojanov2021using} and 2) open-surface dataset Deepfashion3D~\cite{zhu2020deep}).
The test list of ABO, GSO, Meta, and Objaverse is derived from Dora benchmark~\cite{chen2024dora} after excluding our training data which includes about 2.7k assets. For Toys4k, we use the full set following Trellis~\cite{DBLP:journals/corr/abs-2412-01506}.
For the image-to-3D generation, we evaluate the methods on 200 random assets from Toys4k~\cite{stojanov2021using} and some images in the wild, showcasing the superior potential on generation tasks of SparseFlex VAE.

\vspace{-0.4cm}

\paragraph{Baselines.} 
We compare our VAE with previous state-of-the-art methods, including Craftsman~\cite{li2024craftsman}, Trellis~\cite{DBLP:journals/corr/abs-2412-01506}, Dora~\cite{chen2024dora}, XCubes~\cite{ren2024xcube}, Surf-D~\cite{DBLP:conf/eccv/YuDLLLLMKHTLW24} and 3PSDF~\cite{chen20223psdf}, with Surf-D and 3PSDF specially designed for open surfaces.
We directly use the available pre-trained weights provided by these baselines, except for 3PSDF, which we re-implement and train on our dataset.
We only compare Surf-D on the Deepfashion3D dataset due to the lack of available pre-trained weights trained on large-scale datasets.
For the generation results, we compare our method with InstantMesh~\cite{xu2024instantmesh}, Direct3D~\cite{wu2024direct3d}, and TRELLIS~\cite{DBLP:journals/corr/abs-2412-01506}.

\vspace{-0.3cm}

\paragraph{Metrics.} 
We evaluate the reconstruction performance of VAE by using the commonly used metrics, including Chamfer Distance (CD) and F-score with thresholds of 0.01 and 0.001. The metrics are multiplied by $10^4$ and $10^2$, respectively.
For generation results, four orthogonal views of normal maps for each shape are rendered for quantitative comparisons. We report the Fréchet Inception Distance (FID)~\cite{heusel2017gans} and Kernel Inception Distance (KID)~\cite{binkowski2018demystifying}.

\subsection{VAE Reconstruction Evaluation}
We conduct extensive experiments to evaluate the quantitative results of VAE reconstruction from different methods in Table~\cref{tab:vae}.
Since Dora~\cite{chen2024dora} and Craftsman~\cite{li2024craftsman} are trained on watertight data, we separate the results exclusively on the watertight subset from the results computed over the entire dataset for the clear demonstration.
These methods often perform poorly on non-watertight meshes, especially open surfaces like flowers.
Ours$_{256}$ already outperforms other baselines in terms of CD and F-score. 
As the resolution increases, our method achieves even better, ultimately achieving a $\sim$82\% reduction on CD and $\sim$88\% increase in F-score.
\cref{fig:performance_comparison} demonstrate the superiority of our method for complex shapes, open surfaces, and interior structures.
\cref{tab:surfd} compares our method with those designed for open surfaces on the Deepfashion3D~\cite{zhu2020deep} dataset, and our method still achieves the best performance among them.

\begin{figure}
    \centering
    \vspace{-0.6cm}
    \includegraphics[width=1\linewidth]{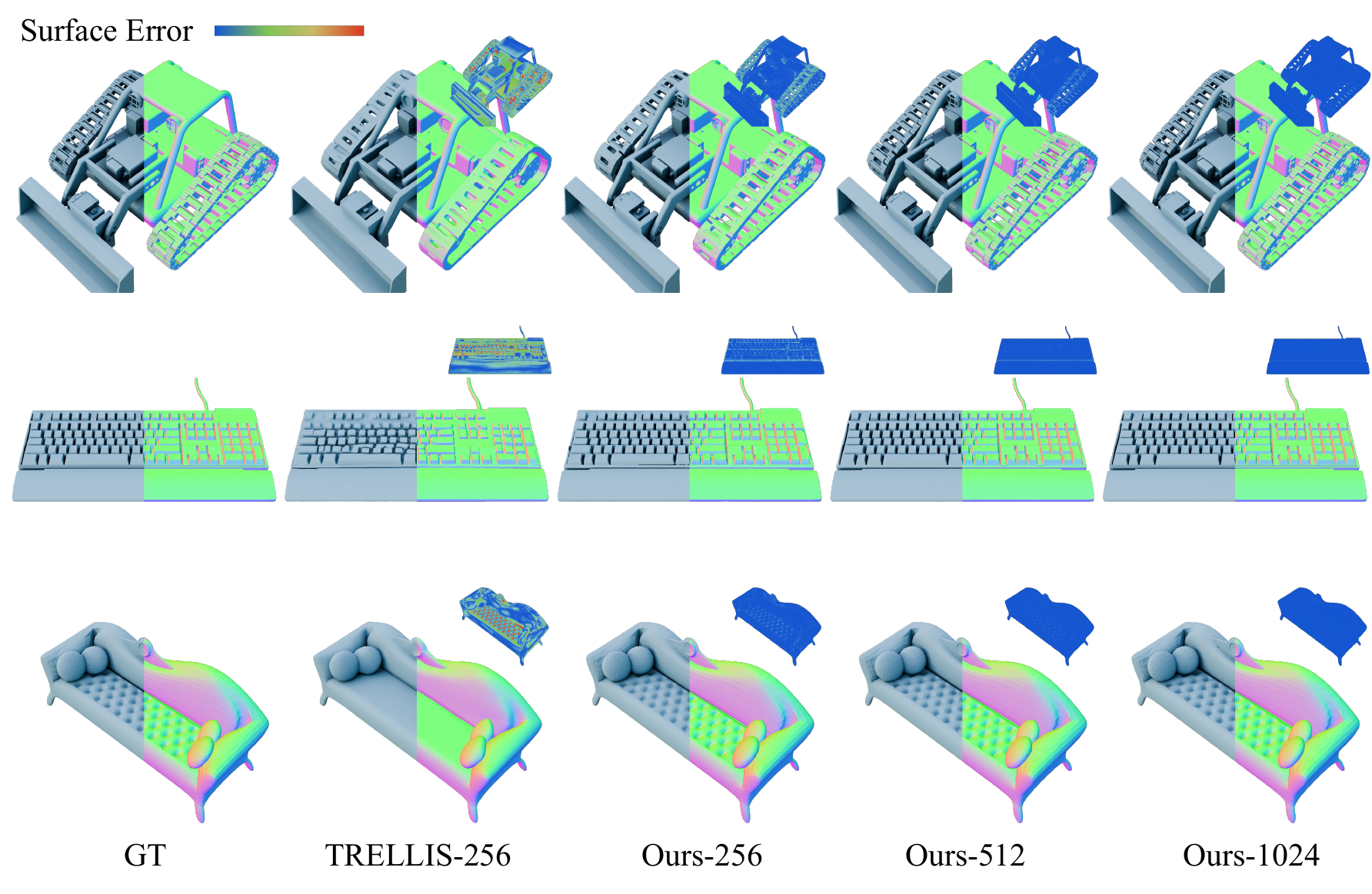}
    \vspace{-0.6cm}
    \caption{\textbf{Qualitative comparison of VAE reconstruction quality} between our method with different resolution and TRELLIS.}
    \label{fig:ab_reso}
    \vspace{-0.2cm}
\end{figure}

\subsection{Image to 3D Genration}
We also validate the effectiveness of our SparseFlex VAE as a foundation model for generation. \cref{tab:generation} validates the effectiveness of our generation. Visualizations are also demonstrated in~\cref{fig:gen_visual}, which includes image-to-3D results using wild images. The generated shapes, which preserves sharp edges and fine details, highly match the given images and showcase the generalization of our method.

\subsection{Ablation Studies}

\input{tables/memory_comparisons}

\input{tables/generation_main}

\paragraph{Self-Pruning Upsampling.}
\cref{tab:surfd} demonstrates that the self-pruning upsampling module plays an important role in the reconstruction quality of open-surface shapes, as it allows for effective pruning of voxels near open boundaries, enabling the reconstruction boundary to better align with input.
Supplementary shows the visual effects of this module in detail.

\vspace{-0.4cm}
\paragraph{\Representation{} and Frustum-aware Sectional Voxel Training.}
To demonstrate the effectiveness of \Representation{} and frustum-aware sectional voxel training strategy, we evaluate the runtime of the feed-forward and GPU memory consumption with different settings, as shown in~\cref{tab:strategy}. 
It demonstrates that \Representation{} effectively reduces the required GPU memory and runtime during network feed-forward. 
However, it is not efficient to scale to higher resolution.
The frustum-aware sectional voxel training strategy eliminates the reliance on the entire surface extraction during rendering, significantly reducing the memory requirements during training.

\begin{figure}
    \centering
    \vspace{-0.6cm}
    \includegraphics[width=1.08\linewidth]{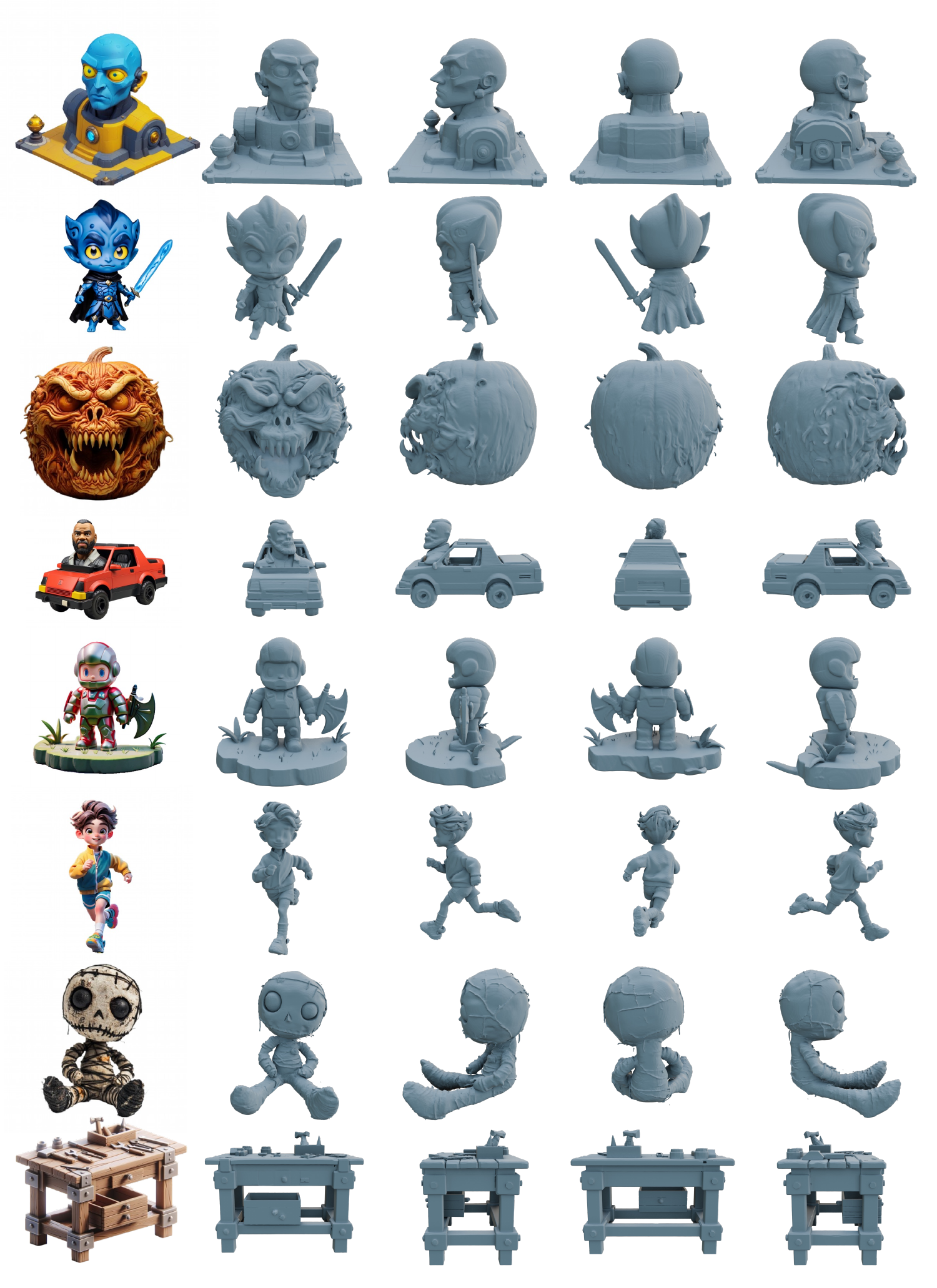}
    \vspace{-0.6cm}
    \caption{\textbf{Single image-to-3D generations with in-the-wild images.}$^1$ The geometry of generated assets accurately preserves sharp edges and fine details.}
    \label{fig:gen_visual}
    \vspace{-0.6cm}
\end{figure}
\vspace{-0.2cm}
\paragraph{Sparse Voxel Resolutions.}
\def\thefootnote{}\footnotetext{$^1$The original images in~\cref{fig:gen_visual} are sourced from various 3D generation platforms, benchmarks (such as Tripo3D, Rodin, Meshy, Trellis, etc.).}
Higher resolution leads to better VAE reconstruction quality as shown in~\cref{tab:vae}.
\cref{fig:ab_reso} illustrates the qualitative comparison of \Representation{} VAE with different resolutions, along with TRELLIS with a resolution of 256.
Thanks to geometry encoding, our approach achieves better geometry reconstruction with the same resolution as TRELLIS.
Benefiting from efficient training, more details of complex structures are revealed, such as the tank track in the first row, as the resolution increases.

%% file: tables/vae_main.tex
\begin{table}[t]
\begin{adjustwidth}{-0.1in}{-0.1in}
\vspace{-0.5cm}
\begin{center}

\footnotesize
\setlength\tabcolsep{2pt}
\resizebox{\linewidth}{!}{
\begin{tabular}{l|ccc|ccc}
\toprule
\multirow{2}{*}{\textbf{Method}} & \multicolumn{3}{c|}{\textbf{Toys4k}} & \multicolumn{3}{c}{\textbf{Dora Benchmark}} \\
\cmidrule(lr){2-4} \cmidrule(lr){5-7}
& $\mathbf{CD\downarrow}$ & $\mathbf{F1(0.001)\uparrow}$ & $\mathbf{F1(0.01)\uparrow}$ & $\mathbf{CD\downarrow}$ & $\mathbf{F1(0.001)\uparrow}$ & $\mathbf{F1(0.01)\uparrow}$ \\
\midrule
Craftsman~\cite{li2024craftsman}         & 13.08/4.63 & 10.13/15.15 & 56.51/85.02 & 13.54/2.06 & 6.30/11.14 & 73.71/91.95 \\
Dora~\cite{chen2024dora}              & 11.15/2.13 & 17.29/26.55 & 81.54/93.84 & 16.61/1.08 & 13.65/25.78 & 78.73/96.40 \\
Trellis~\cite{DBLP:journals/corr/abs-2412-01506}           & 12.90/11.89 & 4.05/4.93 & 59.65/64.05 & 17.42/9.83 & 3.81/6.20 & 62.70/71.95 \\
XCube~\cite{ren2024xcube}             & 4.35/3.14 & 1.61/13.49 & 74.65/79.62 & 4.74/2.37 & 1.31/0.84 & 75.64/86.50 \\
3PSDF$^*$~\cite{chen20223psdf}           & 4.51/3.69 & 11.33/14.10 & 81.70/86.13 & 7.45/1.68 & 7.52/12.50 & 79.43/91.17 \\
\midrule
\rowcolor{pink!10}
Ours$_{256}$               & 2.56/1.25 & 18.31/27.23 & 85.35/92.01 & 1.93/0.53 & 16.24/28.37 & 88.76/97.31 \\
\rowcolor{green!10}
Ours$_{512}$              & 1.67/0.84 & 23.74/34.10 & 90.39/95.60 & 1.36/0.23 & 21.85/36.03 & 91.55/98.51 \\
\rowcolor{blue!10}
Ours$_{1024}$              & \textbf{1.33}/\textbf{0.60} & \textbf{25.95}/\textbf{35.69} & \textbf{92.30}/\textbf{96.22} & \textbf{0.86}/\textbf{0.12} & \textbf{25.71}/\textbf{39.50} & \textbf{94.71}/\textbf{99.14}  \\
\end{tabular}
}
\vspace{-0.2cm}
\caption{\textbf{Quantitative comparison for VAE reconstruction quality on the Toys4K dataset (left) and Dora benchmark (right).} The `/' symbol separates the results computed over the entire dataset from those obtained exclusively on the watertight subset.}
\label{tab:vae}
\vspace{-0.2cm}
\end{center}

\end{adjustwidth}
\end{table}

%% file: tables/deepfashion3d_main.tex
\begin{table}[t]
\begin{center}
\vspace{-0.2cm}

\resizebox{0.65\linewidth}{!}{
\begin{tabular}{l|ccc}
\toprule
\textbf{Method} & $\mathbf{CD\downarrow}$ & $\mathbf{F1(0.001)\uparrow}$ & $\mathbf{F1(0.01)\uparrow}$ \\
\midrule
Surf-D~\cite{DBLP:conf/eccv/YuDLLLLMKHTLW24}            & 63.79 & 0.80 & 23.17 \\
3PSDF$^*$~\cite{chen20223psdf}           & 0.26  & 8.14 & 99.35 \\
\midrule
\rowcolor{pink!10}
Ours$_{256}^\dagger$ & 0.55  & 6.35 & 94.88  \\
\rowcolor{pink!10}
Ours$_{256}$               & 0.08  & 18.60 & 99.99 \\
\rowcolor{green!10}
Ours$_{512}^\dagger$& 0.18 & 11.31 & 99.93 \\
\rowcolor{green!10}
Ours$_{512}$              & \underline{0.05}  & \underline{31.60} & 100.00 \\
\rowcolor{blue!10}
Ours$_{1024}^\dagger$& 0.05 & 24.80 & \underline{100.00} \\
\rowcolor{blue!10}
Ours$_{1024}$              & \textbf{0.04} & \textbf{37.22} & \textbf{100.00} \\
\end{tabular}
}
\vspace{-0.2cm}
\caption{\textbf{Reconstruction results on open-surface dataset Deepfashion3D.} $^\dagger$ indicates the absence of the self-pruning upsampling module.}
\label{tab:surfd}
\vspace{-0.4cm}
\end{center}
\vspace{-0.6cm}

\end{table}

%% file: tables/memory_comparisons.tex
\begin{table}[t]
\begin{adjustwidth}{-0.1in}{-0.05in}
\begin{center}

\resizebox{0.95\linewidth}{!}{
\label{tab:memory}
\begin{tabular}{ccccccc}
\toprule
  & \multicolumn{3}{c}{\textbf{Feed-Forward Time (ms)}$\downarrow$} & \multicolumn{3}{c}{\textbf{GPU Memory Cost (MB)}$\downarrow$} \\
\midrule
Resolution  &  $256^3$  &  $512^3$  &  $1024^3$  &  $256^3$  &  $512^3$  &  $1024^3$  \\
\midrule
Ours ($\alpha=0.1$)  &  333   &  620    &  1151                   &  35515 &  40183  &  55441  \\
Ours ($\alpha=0.3$)   &  357   &  697    &  1475                  &  37403 &  45675  &  69991  \\
w/o FSV      &  390   &  958    &  OOM                       &  40703 &  62029  &  OOM      \\ 
w/o FSV $\&$ Sp.   &  418   &  OOM    &  OOM  &  45505 &  OOM  &  OOM      \\ 
\bottomrule
\end{tabular}
}
\vspace{-0.2cm}
\caption{\textbf{Feed-Forward time and GPU memory cost comparisons.} $\alpha$ stands for the visibility ratio of voxels.
`OOM' means Out Of Memory and `FSV' means frustum-aware sectional voxel training strategy. `Sp' means SparseFlex.}
\label{tab:strategy}
\end{center}
\end{adjustwidth}
\end{table}

%% file: tables/generation_main.tex
\begin{table}[t]
\begin{center}

\vspace{-0.4cm}
\resizebox{\linewidth}{!}{
\begin{tabular}{c|ccc|c}
\toprule
 Method & InstantMesh~\cite{xu2024instantmesh} & Direct3D~\cite{wu2024direct3d} & TRELLIS~\cite{DBLP:journals/corr/abs-2412-01506} &  Ours \\
\midrule
\bf FID$\downarrow$  & 68.74 & 50.84 & 47.66 & \textbf{44.95} \\
\bf KID ($\times 10^3$)$\downarrow$  & 9.68 & 2.04 & 1.28 & \textbf{1.05} \\
\bottomrule
\end{tabular}}
\vspace{-0.2cm}
\caption{\textbf{Quantitative generation results on Toys4k.}}
\label{tab:generation}
\end{center}
\vspace{-0.8cm}

\end{table}

%% file: 10_conclusion.tex
\section{Conclusion}
\label{sec:conclusion}
In this paper, we present \Representation{}, a new sparse-structured isosurface representation for differentiable mesh reconstruction with high resolution using rendering supervision, enabling the reconstruction of open surfaces.
Based on \Representation{}, we propose a novel frustum-aware sectional voxel training strategy with adaptive frustum control to efficiently train \Representation{} VAE with high-resolution, dramatically reducing memory consumption.
This strategy also allows our method to reconstruction the interiors only using rendering loss.
Finally, we develop the image-to-3D generation pipeline following TELLIS~\cite{DBLP:journals/corr/abs-2412-01506}.
Experiments demonstrate state-of-the-art reconstruction accuracy and high-quality generation with open surfaces.

\noindent\textbf{Limitations:} 
Despite the strong performance of SparseFlex VAE in both reconstruction and image-to-3D generation, some limitations remain. 1) Open surface boundaries, while handled effectively by voxel pruning, may exhibit minor artifacts at lower resolutions. 2) High-resolution generation remains computationally demanding. 3) Enhanced control over the generation of interior structures is an area for future work.